\documentclass[10pt,twocolumn,letterpaper]{article}

\usepackage[pagenumbers]{cvpr} % Preprint/arXiv version with visible authors and page numbers

\usepackage{cuted} %< for strip environment

\usepackage{currfile} %< for \currfilebase macro

\usepackage{caption} %< for "\captionof" commands (teaser)
\definecolor{cvprblue}{rgb}{0.21,0.49,0.74}
\usepackage[pagebackref,breaklinks,colorlinks,allcolors=cvprblue]{hyperref}

\def\paperID{*****} % *** Enter the Paper ID here
\def\confName{CVPR}
\def\confYear{2027}

\title{Stable Signer V3: 4D Deformable Primitive Splatting Is All You\\Need for Sign Language Production}

\author{
Sen Fang \qquad Liwei Che \qquad Dimitris N. Metaxas\\
Rutgers University\\
{\hypersetup{urlcolor=magenta}\href{https://signerx.github.io/Stable-Signer-V3}{\texttt{https://signerx.github.io/Stable-Signer-V3}}}
}

\begin{document}
\maketitle
\begin{strip}
\centering
\includegraphics[width=\linewidth]{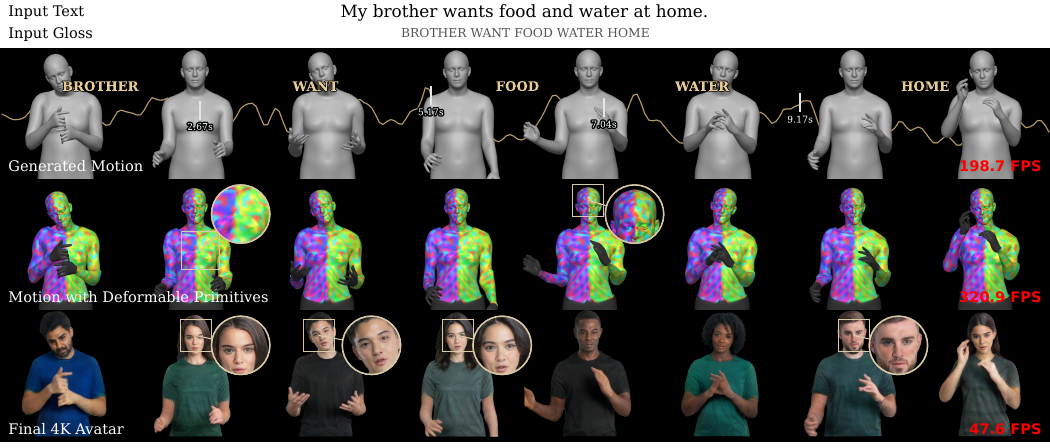}
\captionsetup{hypcap=false}
\captionof{figure}{\textbf{Stable Signer V3.} Given an input sentence or gloss sequence, our system generates signing motion (top), augments the body with deformable face and body primitives (middle), and renders personalized 4K avatars (bottom). Columns show successive moments of the same motion, with four male and four female avatars alternating in the bottom row.}
\captionsetup{hypcap=true}
\label{fig:\currfilebase}%
\end{strip}
\begin{abstract}
We present \textbf{Stable Signer V3}, an interactive high-resolution sign language avatar system built on \textbf{4D Deformable Primitive Splatting}.
It addresses three limitations of existing avatars: restricted facial expression, inaccurate identity-specific body deformation, and loss of fine appearance details.
\textbf{Shared Face Primitive Splatting} adds transferable superquadric primitives for richer expression control; \textbf{Test-Time Body Primitive Splatting} learns an identity-specific virtual muscle layer for more faithful motion; and \textbf{Test-Time Tunneling Super-Resolution} uses TunnelingNet to recover identity-specific high-resolution details.
Both test-time modules are optimized once per signer identity; inference then runs entirely from the stored identity parameters.
Together, they reduce facial geometry error by 71.4\%, body-boundary error by 35.1\% across ten identities, and high-resolution RGB reconstruction error by 24.2\% on held-out evaluations, while retaining interactive streaming.
\end{abstract}    
\section{Introduction}
\label{sec:intro}

%
% 手语生成旨在将文本或词汇序列转化为可理解、自然且具有表现力的手语视频，
% 不同于一般的人体动画，手语的信息不仅由手势传递，还依赖面部表情、头部姿态
% 以及身体运动之间的协调。一个实用的生成系统因此不能仅停留在骨架或参数化
% 姿势的预测：它还需要将这些动作稳定地呈现在具有不同身份与外观的化身上，
% 同时保留细微的表情和视觉细节，并满足交互式应用对响应速度的要求。
Sign language production aims to translate text or lexical sequences into intelligible, natural, and expressive signing videos~\cite{stoll2020text2sign,saunders2020progressive,Fang_2025_ICCV}.
% enabling more direct visual communication for accessibility and human--computer interaction.
Sign language conveys information through coordinated manual gestures, facial expressions, head poses, and body movements~\cite{pfau2010nonmanuals,brentari2018production,saunders2021continuous,10.1145/3706598.3713855}.
A practical production system must therefore go beyond predicting skeletons or parametric poses: it must consistently express these motions through avatars with diverse identities and appearances, preserve subtle expressions and visual details, and meet the latency requirements of interactive applications~\cite{kipp2011assessing,huang2021towards,walsh2025slrtp2025signlanguageproduction}.

% 先前的方法通常来说有Text2Pose2Video和Text2Avatar两种路线，一个通过视频生成模型制作手语视频，一个通过3D模型制作avatar视频。
% 姿势生成模型为手语动作提供了结构化表示，三维高斯泼溅则提供高效的渲染表示，
% 已有工作将其用于手语化身重建和实时上半身动画，
% 但是其中的质量仍然受到限制：
% 能够预测3D姿势和快速渲染，并不意味着已经获得足够精细的手语表达。
% 动作最终如何呈现，仍受到参数化人体的局部表达能力、模板与个体之间的形变匹配，以及图像解码器的细节重建能力的共同限制。
% 因此，我们这篇文章关注的不是重新学习一个完整的语言到视频模型，而是如何在已有动作与化身先验之上，以紧凑的适配机制增强最终的手语表现。
Existing approaches broadly follow two routes.
\textit{Text2Pose2Video} synthesizes signing videos from intermediate poses with pose-conditioned image or video generation models~\cite{stoll2020text2sign,stoll2020signsynth,saunders2020everybodysignnowtranslating,saunders2021anonysign,Saunders_2022_CVPR,SaundersBen20241125,11099440}.
\textit{Text2Avatar} predicts language-conditioned 3D signing motion and uses it to animate explicit characters~\cite{Baltatzis_2024_CVPR,zuo2025signs,low2026signsparkefficientmultilingualsign,hu2024expressive,ivashechkin2025signsplatrenderingsignlanguage,Zhang_2025_ICCV}.
Pose-guided human-video models provide additional tools for this pipeline~\cite{feng2023dreamoving,ma2024followposeposeguidedtexttovideo,zhang2025mimicmotionhighqualityhumanmotion,cheng2025wan}.
Dictionary-based production also exploits reusable signing segments and their transitions~\cite{SaundersBen20241125,walsh-etal-2024-select}.
Pose generation models provide structured representations of signing motion~\cite{saunders2020progressive,saunders2021continuous,xie2023g2pddmgeneratingsignpose,yin2024t2sgptdynamicvectorquantization,Fang_2025_ICCV}.
Within these pipelines, 3D Gaussian splatting~\cite{kerbl3Dgaussians} enables efficient rendering, with recent extensions supporting signing avatars~\cite{hu2024expressive,ivashechkin2025signsplatrenderingsignlanguage} and real-time upper-body animation~\cite{Zhang_2025_ICCV}.
Fine-grained signing remains a central rendering challenge even with accurate 3D pose prediction and fast rendering.
The resulting motion is jointly constrained by the local expressiveness of the parametric human model, the compatibility between template deformation and individual geometry, and the image decoder's ability to reconstruct fine details.
We investigate how compact adaptation mechanisms can improve signing expression by building on existing motion and avatar priors.

% 首先，参数化面部为表情控制提供了稳定接口，但有限的形变空间并不总能充分
% 描述局部的细微运动；仅提高渲染分辨率，无法弥补几何表达的不足。
% 其次，相同的骨架姿态作用于不同化身时，统一的蒙皮与形变规则未必能够贴合
% 每个角色的体型和局部表面运动。即使参考姿态下的外观合理，在动作变化时，
% 这种个体差异仍可能表现为局部形状偏差。
% 最后，紧凑的高斯特征经过通用解码器后，可能丢失身份相关的高频外观细节；
% 简单放大输出图像并不能恢复这些细节。
% 这三个问题分别涉及面部控制、身份相关的身体形变和高分辨率外观，这些问题共同导致了现有方法在手语生成中的表现受限。
Three challenges motivate this work.
First, the compact expression spaces of parametric face models such as FLAME~\cite{FLAME:SiggraphAsia2017} suppress subtle local motion and impose a geometric ceiling on facial detail.
Second, shared skinning and deformation rules cannot consistently accommodate identity-specific body shape and local surface motion~\cite{lei2023gartgaussianarticulatedtemplate,Zielonka2023Drivable3G,gau_human,Pang_2024_CVPR}. An avatar that appears plausible in its reference pose can therefore develop local shape errors as the pose changes.
Third, general-purpose decoders can discard identity-specific high-frequency details when reconstructing images from compact Gaussian features~\cite{Xu_2024_CVPR,Jiang_2024_CVPR,Zhang_2025_ICCV}.
These challenges concern facial control, identity-specific body deformation, and high-resolution appearance, respectively.
% They are difficult to address jointly by merely increasing the number of Gaussians or applying a uniform post-processing stage,
Together, they limit the quality of existing sign language production systems.

% 为此，我们提出 Stable Signer V3，一个基于四维可变形基元泼溅的手语生成系统。
% 核心思路是在已有参数化人体与化身之间引入可学习的局部形变层，
% 并将可以共享的面部运动能力与需要针对身份适配的身体和外观能力分开建模。
% 我们以超二次曲面作为可变形基元的具体实现，使局部残差形变能够随姿势和
% 表情变化，同时保留原有的人体控制接口与化身绑定关系。
% 系统接收文本或 Gloss 输入，使用 SignSparK 完成动作补全，并使用 GUAVA 初始化化身和渲染特征。
To address these challenges, we introduce \textbf{Stable Signer V3}, a sign language production system based on \textbf{4D Deformable Primitive Splatting}.
Our key idea is to introduce a learnable local deformation layer between the parametric human model and the avatar, separating shared facial motion capabilities from identity-specific body and appearance adaptation.
We instantiate the deformable primitives as superquadrics, allowing local residual deformations to vary with pose and expression while preserving the existing human control interface and avatar bindings.
% Given text or gloss input, the system builds on existing pose generation and Gaussian avatar models through three complementary modules.
Given text or gloss input, the system uses SignSparK~\cite{low2026signsparkefficientmultilingualsign} for motion completion and GUAVA~\cite{Zhang_2025_ICCV} for avatar initialization and feature rendering.
A shared facial layer learns local residual deformations driven by FLAME expression, jaw, and eye controls and transfers them to the avatar through surface bindings.
For the body, a pose-conditioned virtual muscle layer is fitted to each identity from reference videos; it corrects geometry before skinning so that bound Gaussians follow the adjusted surface, with displacement and smoothness constraints discouraging implausible deformations.
For appearance, an independently normalized Softmax tunneling path uses terminal decoder semantics to learn identity-specific high-resolution detail while preserving the early Gaussian feature pathway.
Test-time adaptation forms a one-time per-identity character-preparation stage. The resulting body and detail parameters serve every subsequent sequence, cleanly separating identity adaptation from online generation.

Our contributions are summarized as follows:
\begin{itemize}

%
% 【贡献一：通用面部可变形基元】
% 通用面部基元泼溅在 FLAME 控制空间之上学习局部形变残差，
% 由表情、下颌、眼部等控制参数驱动，使原有面部表示获得额外的表达自由度。
% 我们学习的是可变形基元及其控制关系，而不是为每个角色重新训练一组高斯。
% 该模块作为共享的面部增强层，再通过与面部表面的绑定传递形变到化身。
\item \textbf{Shared Face Primitive Splatting.} A shared deformable primitive layer enables more complex and controllable facial expressions.

%
% 【贡献二：测试时身体适配】
% 测试时身体基元泼溅利用目标角色的参考视频，学习角色专属、姿势条件化的
% “虚拟肌肉”层。它在蒙皮之前修正基础人体的局部形状，使既有绑定的高斯
% 随修正后的表面运动，而不依赖重新学习整个化身的外观。
% 我们以视频中的投影证据监督适配，并约束形变幅度与空间平滑性，
% 在拟合角色特征的同时限制不合理的局部变形。
\item \textbf{Test-Time Body Primitive Splatting.} An identity-specific virtual muscle layer adapts body deformation from reference videos while keeping the avatar generation backbone frozen.

%
% 【贡献三：测试时隧穿细节增强】
% 测试时隧穿超分辨率利用角色参考视频中的高分辨率信息，
% 在预训练渲染解码器的末端引入独立归一化的 Softmax 隧穿路径。
% 该路径利用已有的图像生成能力学习身份相关的细节补偿，而不是在尚未形成图像语义的早期高斯特征中直接混入视频输入。
% 身体适配与细节增强均在角色准备阶段进行一次测试时优化；
% 后续生成复用已学习的参数，不必在每次推理时重新提供参考视频或进行优化。
% 这使我们能够将角色专属的适配成本与在线手语生成过程分离，
% 在保留已有控制流程的基础上，面向稳定、精细且高效的高分辨率化身表达。
\item \textbf{Test-Time Tunneling Super-Resolution.} A late tunneling path learns identity-specific high-resolution detail from reference videos and stores it in the deployed decoder.
\end{itemize}

\section{Related Work}
\label{sec:related_work}

% 手语生成。
% 手语生成通常有两条路线：一条先从文本或 Gloss 生成姿势，再通过姿势条件驱动的视频
% 生成模型制作手语视频；另一条用生成的动作驱动三维人体或化身，直接渲染视频。
% 在动作生成上，一类方法从已有词典中检索并拼接手语片段，另一类方法训练模型
% 根据语言输入预测连续动作。SignSparK 结合稀疏关键帧控制与条件流匹配，
% 生成稠密的三维手语动作。围绕这些路线，已有工作分别研究了动作准确性、
% 片段之间的连贯性、人物外观和生成速度。我们的系统复用这一动作补全能力，
% 重点研究个性化化身渲染所需的面部、身体与外观适配。

\noindent\textbf{Sign Language Production.}
Resources such as WLASL~\cite{li2020word}, How2Sign~\cite{duarte2021how2sign}, JWSign~\cite{gueuwou2023jwsign}, and YouTube-SL-25~\cite{Tanzer2024YouTubeSL25AL,fang2026signverse2mtwomillionclipposenativeuniverse,fang2026signxcontinuoussignrecognition} support research across lexical signs, continuous signing, and multiple languages.
Pose-conditioned video synthesis has been explored through Text2Sign~\cite{stoll2020text2sign}, SignSynth~\cite{stoll2020signsynth}, and Everybody Sign Now~\cite{saunders2020everybodysignnowtranslating}, with subsequent work on appearance control, large-scale co-articulation, and diffusion-based synthesis~\cite{saunders2021anonysign,Saunders_2022_CVPR,11099440,fang-etal-2026-stable}.
Broader human-video synthesis develops pose-guided generation through DreaMoving~\cite{feng2023dreamoving}, Follow Your Pose~\cite{ma2024followposeposeguidedtexttovideo}, and MimicMotion~\cite{zhang2025mimicmotionhighqualityhumanmotion}, with ControlVideo~\cite{zhang2023controlvideotrainingfreecontrollabletexttovideo} and Wan-Animate~\cite{cheng2025wan} further exploring controllable video generation and character animation.
Multimodal research spans video modeling~\cite{shen2026chronovision,quan2026reinforcingconsistencyvideomllms,shen2026egoforge}, information flow~\cite{zhang-etal-2025-redundancy}, and visual attention~\cite{zhang-etal-2025-shallow}.
The 3D route combines signing-motion generation~\cite{Baltatzis_2024_CVPR,zuo2025signs,low2026signsparkefficientmultilingualsign} with controllable avatar rendering~\cite{hu2024expressive,ivashechkin2025signsplatrenderingsignlanguage,Zhang_2025_ICCV}.
For motion generation, some methods retrieve and concatenate signing clips from existing dictionaries~\cite{zelinka2020neural,Saunders_2022_CVPR,SaundersBen20241125,walsh-etal-2024-select}, while others learn to predict continuous motion from language inputs using progressive prediction or motion primitives~\cite{saunders2020progressive,saunders2021continuous,SLP_Mixed}.
Motion modeling has also explored adversarial training, skeletal graph structure, and non-autoregressive generation~\cite{saunders2020adversarial,saunders2021skeletal,hwang2021non}.
Text-to-human-motion research explores diffusion, language-like motion modeling, and fine-grained gait behavior~\cite{zhang2022motiondiffusetextdrivenhumanmotion,tevet2022humanmotiondiffusionmodel,jiang2023motiongpt,shen2026decoding}.
Large language models are increasingly applied to sign-language translation, understanding, and production~\cite{gong2024llms,inan-etal-2025-signalignlm,Fang_2025_ICCV}.
Recent motion generators explore discrete diffusion for gloss-to-pose synthesis~\cite{xie2023g2pddmgeneratingsignpose}, autoregressive motion tokens~\cite{yin2024t2sgptdynamicvectorquantization}, and multilingual generation from text and prompts~\cite{Fang_2025_ICCV}.
SignSparK~\cite{low2026signsparkefficientmultilingualsign,fang2025streamflowtheoryalgorithmimplementation,fang2026racrectifiedflowauto} combines sparse keyframe control with conditional flow matching to generate dense 3D signing motion.
Appearance transfer in skeletal pose sequences further studies how signer-specific characteristics can be changed while retaining signing motion~\cite{moryossef2025posebasedsignlanguageappearance}.
Together, these directions address motion accuracy, co-articulation, appearance control, and efficient generation~\cite{huang2021towards,walsh2025slrtp2025signlanguageproduction}.

% 用于手语的三维人体与高斯化身。
% SMPL-X 提供身体、手和面部的统一控制，FLAME 提供面部形状、表情与关节控制。
% 这些参数化模型为三维手语化身提供控制接口，而非手部信息的表达也依赖面部
% 和身体动作的准确呈现。在三维高斯泼溅的基础上，SplattingAvatar 将高斯绑定
% 到可变形网格，GART 使用可学习蒙皮和潜在骨骼表示非刚性运动，
% SignSplat 则通过序列级网格拟合和高斯正则化改善手语中的手部与面部重建。
% 我们保留结构化控制与表面绑定，加入共享面部残差和身份相关的身体形变。

\noindent\textbf{3D Humans and Gaussian Avatars for Sign Language.}
SMPL-X~\cite{Pavlakos_2019_CVPR} provides joint body, hand, and facial controls for animatable humans, while FLAME~\cite{FLAME:SiggraphAsia2017} models facial shape, expression, and articulation.
These parametric representations support pose-driven signing avatars~\cite{Baltatzis_2024_CVPR,zuo2025signs}.
Whole-body reconstruction methods, including SMPLer-X and component-aware mesh recovery~\cite{cai2023smplerx,lin2023one}, and dedicated hand reconstruction~\cite{pavlakos2024reconstructing} support detailed avatar controls; DWPose provides complementary whole-body keypoint estimates~\cite{yang2023effective}.
Complementary work improves 3D pose estimation for sign language~\cite{ivashechkin_sltat} and resolves hand-to-hand intersections~\cite{ivashechkin_two_hands}, supporting more reliable motion inputs.
Accurate facial and body motion is particularly relevant to signing, where non-manual cues contribute to meaning~\cite{kipp2011assessing,10.1145/3706598.3713855}.
Building on 3D Gaussian splatting~\cite{kerbl3Dgaussians}, SplattingAvatar~\cite{SplattingAvatar:CVPR2024} embeds Gaussians on a deformable mesh surface, while GART~\cite{lei2023gartgaussianarticulatedtemplate} learns forward skinning and latent bones to capture articulated and non-rigid motion.
Articulated Gaussian models further explore cage-based deformation in D3GA~\cite{Zielonka2023Drivable3G}, monocular human reconstruction in GauHuman~\cite{gau_human}, and efficient human rendering in ASH and HiFi4G~\cite{Pang_2024_CVPR,Jiang_2024_CVPR}.
Related efforts address monocular Gaussian avatar reconstruction in GVA~\cite{liu2024gvareconstructingvivid3d}, image-based body estimation in GST~\cite{prospero2024gstprecise3dhuman}, and generalizable sparse-view synthesis in GHG~\cite{kwon2024ghg}.
Expressive avatar reconstruction is also studied by X-Avatar~\cite{shen2023xavatar} and EVA~\cite{hu2024expressive}, with EVA learning Gaussian human avatars from monocular RGB video.
SignSplat~\cite{ivashechkin2025signsplatrenderingsignlanguage} adapts Gaussian avatars to signing through sequence-level mesh fitting and Gaussian regularization, emphasizing accurate hands and facial expressions.

% 个性化化身与测试时训练。
% 逐身份优化能够重建角色细节，例如 Gaussian Head Avatar 联合学习中性高斯
% 和表情驱动的形变场，实现高质量的 2K 头部渲染。参考图像条件下的图像与
% 视频合成也可以迁移人物外观。对于三维化身，GUAVA 从单图预测可驱动的
% 上半身高斯化身，并通过神经解码器细化渲染特征。
% 我们将通用化身初始化与逐身份适配结合，在角色准备阶段利用参考视频训练
% 身体适配层和末端解码器细节模块。预训练主干保持冻结，适配仅执行一次；
% 在线手语生成时复用所得参数，无需参考输入或再次优化。

\noindent\textbf{Personalized Avatars and Test-Time Training.}
Video-based character reconstruction has developed through HumanNeRF~\cite{weng_humannerf_2022_cvpr} and NeuMan~\cite{jiang2022neuman}, with Vid2Avatar exploring self-supervised scene decomposition for videos in the wild~\cite{guo2023vid2avatar}.
Efficient character preparation is explored by InstantAvatar~\cite{jiang2022instantavatar}, which learns avatars from monocular video.
Subject-specific optimization can recover detailed avatar appearance: Gaussian Head Avatar~\cite{Xu_2024_CVPR} jointly learns neutral Gaussians and an expression-driven deformation field for high-fidelity 2K head rendering.
Reference-conditioned image and video synthesis transfers appearance from exemplar images~\cite{zakharov2020fast,wang2019few}.
High-resolution conditional synthesis and latent diffusion provide complementary image-generation frameworks~\cite{wang2018pix2pixHD,Rombach_2022_CVPR}, with ControlNet~\cite{zhang2023adding} and ControlNeXt~\cite{peng2025controlnextpowerfulefficientcontrol} exploring additional conditioning for image and video generation.
For 3D avatars, GUAVA~\cite{Zhang_2025_ICCV} predicts an animatable upper-body Gaussian avatar from a single image and refines rendered features with a neural decoder.
Lightweight adaptation offers a way to reuse pretrained representations: LoRA~\cite{Hu-ICLR-2022-LoRA} learns low-rank weight updates, and LLaMA-Adapter~\cite{Zhang-arXiv-2023-LLaMA-Adapter} introduces new conditioning through zero-initialized attention.
Adapters also connect pretrained visual representations to dense prediction tasks~\cite{chen2022vision}.
% Broader multimodal cognition, federated learning, and MLLM reliability references
% are retained in the bibliography but omitted here to keep the discussion focused
% on avatar personalization and test-time adaptation.
Character reference videos supervise a body deformation layer and a detail module at the decoder's terminal stage, with the pretrained backbone held fixed.
Both modules are optimized once per signer, and online signing directly reuses the stored parameters.

% Polished vector diagram with illustrative avatar renderings.
% Figure 2 V8. Editable source: make/method_figures/fig2_overview.drawio.
\begin{figure*}[t]
    \centering
    \includegraphics[width=\textwidth]{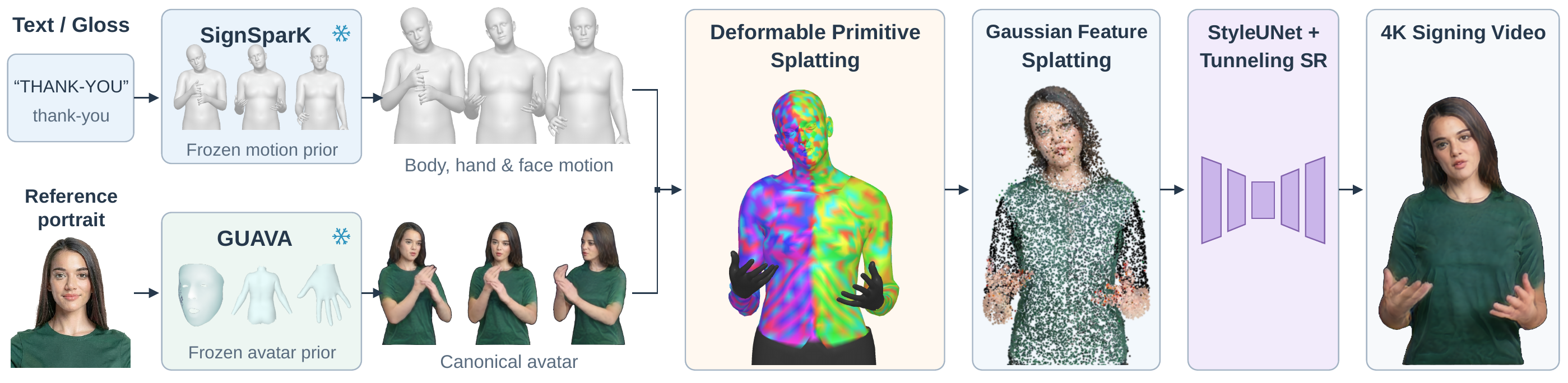}
    \caption{\textbf{Overview of Stable Signer V3.} Motion controls drive an avatar through deformable primitives. Gaussian feature splatting and a tunneling-enhanced decoder produce high-resolution signing videos.}
    \label{fig:method}
\end{figure*}
\section{Method}
\label{sec:method}

\subsection{Overview}
Given text or gloss input, SignSparK~\cite{low2026signsparkefficientmultilingualsign} completes dictionary-retrieved sparse poses into body, hand, and facial controls $\{\mathbf{q}_t\}_{t=1}^{T}$. GUAVA~\cite{Zhang_2025_ICCV} initializes a surface-bound Gaussian avatar from a reference image. We augment this pipeline with shared facial deformation and identity-specific body and appearance adaptation.

For signer $s$, aligned reference images, controls, and cameras $\mathcal{V}_s=\{I_t,\mathbf{q}_t,\pi_t\}$ supervise image formation:
\begin{equation}
 \widehat I_t=\mathcal{D}_{\psi_s}\!\left(
 \mathcal{R}_{\pi_t}\!\left[\mathcal{G}_s(\mathbf{q}_t;\phi,\theta_s)\right]\right),
 \label{eq:pipeline}
\end{equation}
where $\phi$ is the shared facial controller, $\theta_s$ is the body adaptation, and $\mathcal{G}_s$ transports Gaussians with the deformed surface. The rasterizer $\mathcal{R}$ projects Gaussian features, and $\mathcal{D}_{\psi_s}$ decodes them into high-resolution RGB. Adaptation occurs once per identity; inference runs from stored parameters.

\subsection{Deformable Primitive Representation}
\label{sec:primitive_representation}
Here \emph{4D} denotes pose- and time-conditioned primitive transformations across motion sequences.
A sparse field blends control-dependent affine residuals over superquadric supports. Primitive $k$ has center $\mathbf{c}_k$, orthonormal frame $\mathbf{R}_k$, positive axis lengths $\mathbf{a}_k$, and normalized distance
\begin{equation}
 d_k(\mathbf{x})=\left\|
 \operatorname{diag}(\mathbf{a}_k)^{-1}
 \mathbf{R}_k^{\mathsf T}(\mathbf{x}-\mathbf{c}_k)\right\|_4.
 \label{eq:sq_support}
\end{equation}
The level set $d_k(\mathbf{x})=1$ defines a superellipsoid with fixed shape exponents. Centers are initialized in bilateral pairs, and local surface covariance determines orientations and scales. The supports remain fixed during controller learning. Each surface sample receives a sparse neighborhood $\mathcal{N}(\mathbf{x})$ with weights
\begin{equation}
 w_k(\mathbf{x})=\frac{\exp[-2d_k(\mathbf{x})^2]}
 {\sum_{j\in\mathcal{N}(\mathbf{x})}\exp[-2d_j(\mathbf{x})^2]}.
 \label{eq:binding}
\end{equation}
Fixed reference-surface bindings blend controller-predicted affine residuals $(\mathbf{A}_{k,t},\mathbf{b}_{k,t})$:
\begin{equation}
 \Delta\mathbf{x}_t=\sum_{k\in\mathcal{N}(\mathbf{x})}w_k(\mathbf{x})
 \left[\mathbf{A}_{k,t}(\mathbf{x}_t-\mathbf{c}_k)+\mathbf{b}_{k,t}\right],
 \label{eq:primitive_field}
\end{equation}
Here $\mathbf{x}_t$ uses local coordinates. Zero-initialized controller outputs preserve geometry. The field deforms the surface, while bound Gaussians supply appearance.

% Figure 3 V10 component diagram; previous versions are retained locally.
\begin{figure*}[t]
    \centering
    \includegraphics[width=\textwidth]{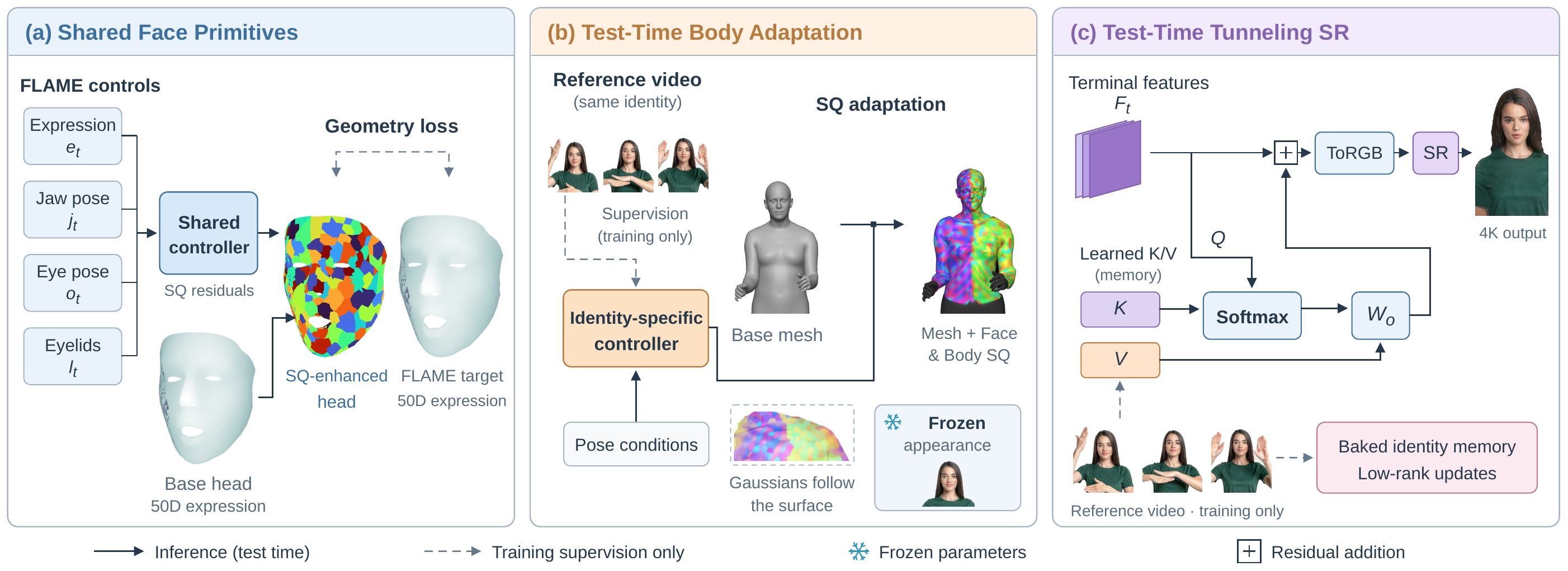}
    \caption{\textbf{Deformation and detail adaptation.} (a) Shared facial primitives learn geometric residuals with FLAME supervision. (b) Identity-specific body primitives adapt the avatar carrier. (c) A late Softmax tunneling pathway learns high-resolution appearance. Dashed arrows denote training supervision; solid arrows denote inference.}
    \label{fig:method_components}
\end{figure*}

\subsection{Shared Face Primitive Splatting}
\label{sec:face_primitives}
The shared facial field is conditioned on
\begin{equation}
 \mathbf{u}^{\mathrm{face}}_t=[\mathbf{e}_t,\mathbf{j}_t,\mathbf{o}_t,\boldsymbol{\ell}_t],
\end{equation}
comprising 50 expression coefficients, three jaw parameters, six eye-pose parameters, and two eyelid controls. Global head orientation is handled by the body rig. A multilayer perceptron maps normalized facial controls to the affine residuals in Eq.~\ref{eq:primitive_field}. Common facial surface correspondences provide the interface for transferring the shared controller across identities.

The FLAME baseline and FLAME+SQ use the same fifty expression coefficients together with identical jaw, eye, and eyelid controls. Registered target geometry supervises only the SQ residual. For corresponding surface samples $\mathbf{x}_{ti}$ and $\mathbf{y}_{ti}$, we minimize
\begin{equation}
\begin{split}
 \mathcal{L}_{\mathrm{face}}={}&\frac{1}{|\Omega|}\sum_{(t,i)\in\Omega}
 \omega_i\rho(\mathbf{x}_{ti}+\Delta\mathbf{x}_{ti}-\mathbf{y}_{ti})\\
 &+\lambda_f\mathbb{E}_t\|\mathbf{T}_t\|_F^2,
\end{split}
 \label{eq:face_loss}
\end{equation}
where $\rho$ is a smooth $L_1$ penalty, $\Omega$ indexes training samples, and $\mathbf{T}_t$ collects affine coefficients. Weights $\omega_i$ emphasize lips and eyes. Appearance, opacity, and the decoder remain frozen; compatible facial bindings transfer the controller across identities.

\subsection{Test-Time Body Primitive Splatting}
\label{sec:body_primitives}
An identity-specific virtual muscle layer covers the torso, shoulders, arms, and wrists. Static and pose-dependent affine corrections capture individual surface motion:
\begin{equation}
 \mathbf{T}_{s,t}=\tanh\!\left(\mathbf{T}^{\mathrm{static}}_s+
 f_{\theta_s}(\overline{\mathbf{p}}_t)\right),
 \label{eq:body_controller}
\end{equation}
where $\overline{\mathbf{p}}_t$ is the normalized 63-dimensional body pose. Static and dynamic components capture shape and articulation effects, with bounded coefficients limiting distortion.

The blended field modifies the canonical carrier before skinning:
\begin{equation}
 \widetilde{\mathbf{V}}_{s,t}=\operatorname{LBS}\!\left(
 \mathbf{V}^{\mathrm{can}}_s+\mathcal{S}(\Delta\mathbf{V}_{s,t}),\mathbf{q}_t\right).
 \label{eq:body_skinning}
\end{equation}
Here $\mathcal{S}$ smooths offsets along body mesh edges, preserving the head, hands, and motion controls. Template and UV Gaussians follow corrected vertices and parent triangles, respectively.

Differentiable rendering supervises a dilated visibility region of body-bound Gaussians. Raw and decoded RGB, gradients, SSIM~\cite{wang2004image}, silhouette, and boundary losses enforce video alignment:
\begin{equation}
\begin{split}
 \mathcal{L}_{\mathrm{body}}={}&\lambda_r\mathcal{L}_{\mathrm{raw}}+
 \lambda_c\mathcal{L}_{\mathrm{rgb}}+\lambda_g\mathcal{L}_{\mathrm{grad}}+
 \lambda_s\mathcal{L}_{\mathrm{ssim}}\\
 &+\lambda_m\mathcal{L}_{\mathrm{mask}}+
 \lambda_b\mathcal{L}_{\mathrm{boundary}}+\mathcal{L}_{\mathrm{reg}}.
\end{split}
 \label{eq:body_loss}
\end{equation}
Regularization penalizes affine magnitude, neighboring translations, and mean vertex displacement. The avatar generator, appearance, and renderer remain frozen. Held-out reconstruction guides selection.

\subsection{Test-Time Tunneling Super-Resolution}
\label{sec:detail_tunneling}
A late Softmax tunnel adapts terminal decoder features into identity-specific detail before RGB prediction.

Let $\mathbf{h}_t(\mathbf{p})$ be the terminal feature at location $\mathbf{p}$. Learnable spatial keys and values $\{\mathbf{k}_{s,n},\mathbf{v}_{s,n}\}_{n=1}^{N}$ are initialized from reference frames and optimized by video reconstruction. Interpolation to the feature resolution gives
\begin{align}
 \alpha_{t,n}(\mathbf{p})&=\frac{\exp[\mathbf{z}_t(\mathbf{p})^{\mathsf T}\mathbf{k}_{s,n}(\mathbf{p})/\sqrt{r}]}
 {\sum_{m=1}^{N}\exp[\mathbf{z}_t(\mathbf{p})^{\mathsf T}\mathbf{k}_{s,m}(\mathbf{p})/\sqrt{r}]},\\
 \widetilde{\mathbf{h}}_t(\mathbf{p})&=\mathbf{h}_t(\mathbf{p})+
 W_o\sum_{n=1}^{N}\alpha_{t,n}(\mathbf{p})\mathbf{v}_{s,n}(\mathbf{p}),
 \label{eq:tunnel}
\end{align}
Here $N$ is the number of memory slots, $\mathbf{z}_t(\mathbf{p})=W_q\mathbf{h}_t(\mathbf{p})$, and $r$ is the query/key dimension. Projections $W_q$ and $W_o$ map into and out of the tunnel, while $\alpha_{t,n}$ weights slot $n$. Softmax normalizes slots per location; zero-initialized $W_o$ preserves the original feature path.

Low-rank updates $\Delta W=BA$, with trainable factors $A$ and $B$, complement the tunnel with pretrained weights frozen. Training and inference use one StyleUNet~\cite{Wang_2021_CVPR} pass; stored identity memory supplies the deployed inference path.

\paragraph{High-resolution reconstruction.}
Let $J_t$ be the decoded RGB image and $\mathcal{U}_q$ denote spatial upsampling by factor $q$. The intermediate reconstruction is
\begin{equation}
 \widehat I_t^{\mathrm{mid}}=\mathcal{U}_q(J_t)
 +\mathcal{H}_s(\widetilde{\mathbf h}_t,J_t),
 \label{eq:sr_residual}
\end{equation}
where $\mathcal{H}_s$ is an identity-adapted PixelShuffle residual head. A shared fully convolutional head refines aligned native-resolution crops for 4K output. High-resolution target patches preserve supervision detail.

\paragraph{Appearance objective.}
We learn appearance parameters from aligned video frames with fixed geometry and Gaussian features. Intermediate-resolution supervision combines RGB, image-gradient, multiscale high-pass, detail-band, local-contrast, perceptual, and low-frequency penalties. The native-resolution head receives RGB, gradient, high-pass, and local-detail supervision:
\begin{equation}
 \mathcal{L}_{\mathrm{detail}}=\lambda_{\mathrm{mid}}\mathcal{L}_{\mathrm{mid}}
 +\mathbb{E}_{t,\mathcal{P}}\sum_a\lambda_a
 \mathcal{L}_a(\widehat I_t^{\mathcal{P}},I_t^{\mathcal{P}}),
 \label{eq:detail_loss}
\end{equation}
Here $\mathcal{L}_{\mathrm{mid}}$ is the intermediate loss above; $\lambda_{\mathrm{mid}}$ and $\lambda_a$ are nonnegative weights. The crop $\mathcal{P}$ is shared by prediction and target, and $a$ indexes the native-resolution loss terms. Patch sampling preserves target-frequency information while bounding training memory. Body and detail adaptation are optimized with their respective frozen components, yielding a geometric controller and an appearance decoder for each identity. Online signing combines these modules with the shared facial field and newly generated motion.

\begin{figure}[t]
\centering
\includegraphics[width=\linewidth]{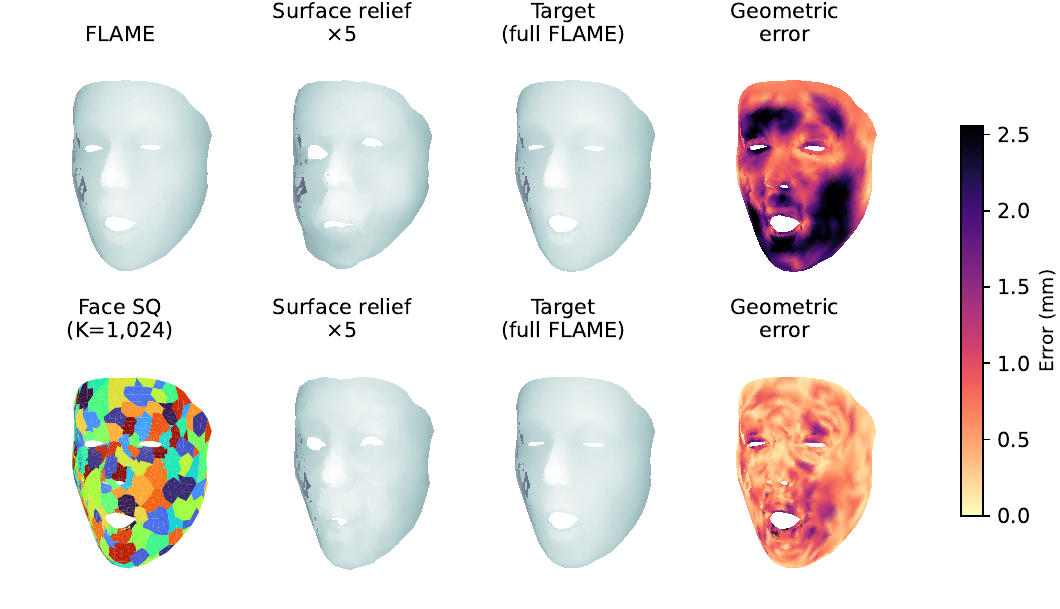}
\caption{\textbf{Face primitive reconstruction on a held-out frame.} Top: full-expression FLAME. Bottom: FLAME+SQ geometry, colored by primitive assignment. The second column amplifies surface deviations from the target by $5\times$ for visibility. The third column shows the registered target; the last shows unamplified geometric error on a shared scale (darker indicates larger error).}
\label{fig:ph_face}
\end{figure}

% Declare the body float on page 6 for placement at the top of page 7.
\begin{figure*}[t]
\centering
\includegraphics[width=\textwidth]{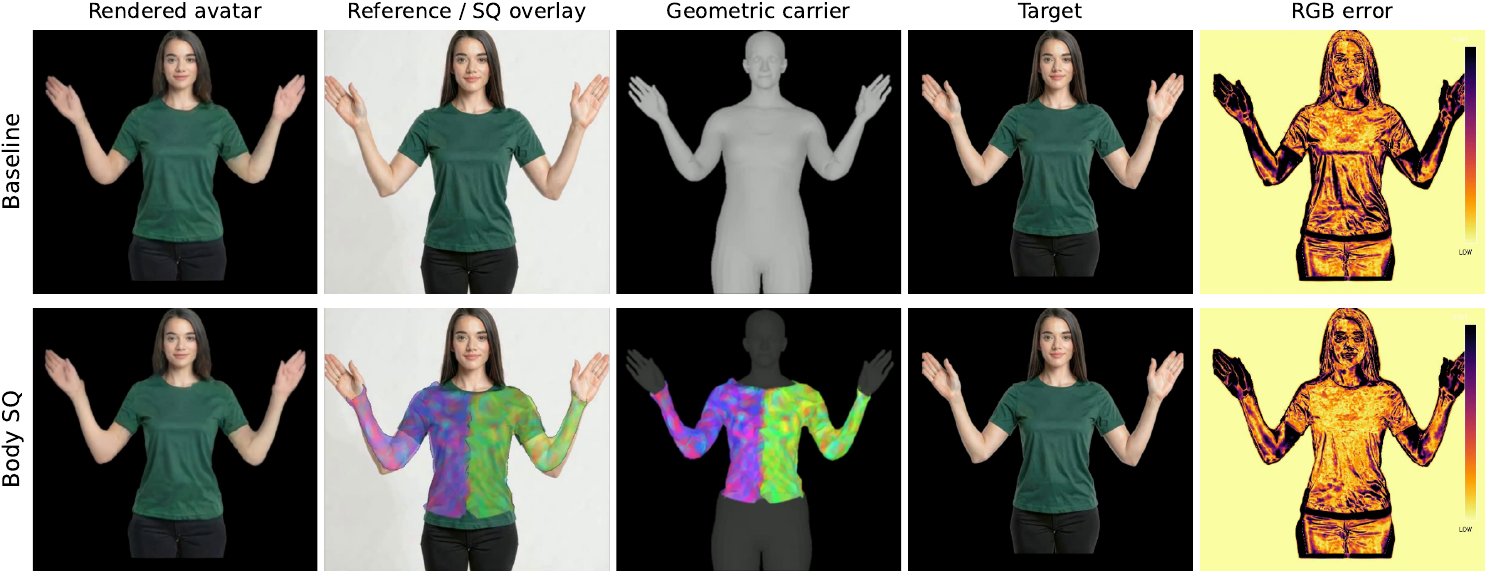}
\caption{\textbf{Test-time body adaptation.} Frozen GUAVA (top) and identity-specific Body SQ (bottom). Columns show the rendered avatar, reference frame with adapted SQ patches overlaid at 60\% opacity, geometric carrier, target, and RGB error on a shared scale. The top-row reference has no SQ overlay. The comparison clearly shows that Body SQ captures large muscle-driven arm deformations and brings the 3D carrier into closer alignment with the signer, producing a more stable body under demanding motion.}
\label{fig:body_sq}
\end{figure*}

\section{Experiments}
\label{sec:experiments}
% Three-page budget: three component studies with qualitative evidence,
% followed by one overall quantitative comparison. Extra studies stay optional.
% Imported figures must be replaced before submission; retain source attribution while displayed.

\subsection{Experimental Setup}
The evaluation covers facial expression reconstruction, identity-specific body adaptation, and high-resolution detail recovery.

\noindent\textbf{Data and protocol.} Controlled face and body studies use chronological 80\%/20\% training--evaluation splits and fixed optimization budgets; checkpoint selection uses the training segment exclusively. Detail adaptation is evaluated on held-out frames, and cross-method rendering uses identical poses, cameras, backgrounds, and target-defined crops for every method. Scores are first aggregated per identity and then macro-averaged.

\noindent\textbf{Metrics.} Facial geometry is measured by Euclidean point error in the native shared tracked coordinates. Body adaptation reports refined RGB L1, boundary L1, and SSIM~\cite{wang2004image}. Rendering comparisons use foreground PSNR~\cite{4775883} together with SSIM and LPIPS~\cite{Zhang_2018_CVPR} inside a shared ground-truth person crop. Runtime measurements distinguish independently warmed stage throughput from measured resident-pipeline wall time. Semantic preservation is evaluated through fixed back-translation using BLEU-4~\cite{papineni2002bleu} and ROUGE-L F1~\cite{lin-2004-rouge} on poses re-extracted from the rendered RGB videos.

\subsection{Ablation I: Shared Face Primitives}
We train the shared face layer on registered facial geometry, using a chronological 80\%/20\% training--evaluation split and a fixed optimization budget. Full-expression FLAME and FLAME+SQ receive identical fifty-dimensional expression and articulation controls; only the SQ residual distinguishes the two configurations.

Table~\ref{tab:main_geometry} reports Euclidean error over all retained head samples, including template- and UV-bound samples: mean head error falls from 1.267 to 0.362\,mm (71.4\%), and facial error falls from 1.492 to 0.518\,mm. Figure~\ref{fig:ph_face} shows the held-out midpoint with a shared error scale and $5\times$ deviation magnification. The shared primitive controller reduces head error by 71.4\% and recovers fine expression geometry across the face; region-wise results are provided in the supplement.
% Recomputed from archived checkpoint; see make/ablation_results_v1/face_evaluation.json.
% Share the natural width of Table 1 with Table 2.
\newsavebox{\geometryTableBox}
\newlength{\comparisonTableWidth}
\begin{table}[t]
\centering\small
\setlength{\tabcolsep}{4pt}
\begin{lrbox}{\geometryTableBox}
\begin{tabular}{l|rrr}
\hline
Method & Head & Face & Lips\\\hline
FLAME & 1.267 & 1.492 & 1.739 \\
FLAME+SQ & 0.362 & 0.518 & 0.990 \\
Reduction (\%) & 71.4 & 65.3 & 43.1 \\
\hline
\end{tabular}
\end{lrbox}
\global\comparisonTableWidth=\wd\geometryTableBox
\usebox{\geometryTableBox}
\typeout{Comparison table width: \the\comparisonTableWidth}
\caption{\textbf{Face reconstruction error (mm, lower is better).} Average over the held-out 20\% of the sequence. Head aggregates all retained samples; face and lip metrics use their semantic subsets. Registered facial surfaces provide the target geometry, and both methods receive identical controls.}
\label{tab:main_geometry}
\end{table}

% Matched historical Body SQ runs with the same adaptation budget and split.
% Provenance: make/ablation_results_v1/body_capacity_evaluation.json.
\begin{table}[t]
\centering\small
\setlength{\tabcolsep}{4pt}
\resizebox{\comparisonTableWidth}{!}{%
\begin{tabular}{l|rrr}
\hline
Body model & RGB L1 $\downarrow$ & Boundary L1 $\downarrow$ & SSIM $\uparrow$\\\hline
Baseline & 0.0649 & 0.0350 & 0.9038\\
$K=512$ & 0.0599 & 0.0282 & 0.9076\\
$K=1{,}024$ & 0.0575 & 0.0270 & 0.9096\\
\hline
\end{tabular}}
\caption{\textbf{Body primitive count.} Baseline and Body SQ use the same chronological 80\%/20\% train--evaluation split for one identity. Both SQ variants share the adaptation budget and loss weights. Metrics are computed in the body region; RGB L1 uses refined renderings.}
\label{tab:body_capacity}
\end{table}

% Use the table float queue to retain Figure 6 beside the page-6 tables,
% independently of the earlier full-width Figure 5 scheduled for page 7.
\begin{table}[t]
\centering
\includegraphics[width=\linewidth]{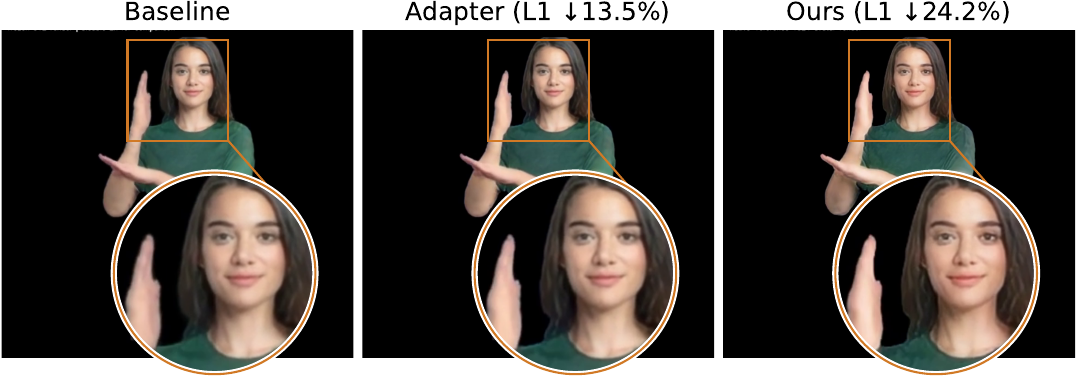}
\captionof{figure}{\textbf{Detail adaptation.} Frozen baseline, Adapter, and our tunneling decoder at the same timestamp. Circular insets magnify facial regions. The baseline is enlarged from 512 resolution; both outputs are generated at 1024. Parentheses report mean held-out RGB L1 reductions relative to the frozen baseline.}
\label{fig:ph_rendering}
\end{table}

\subsection{Ablation II: Test-Time Detail Tunneling}
Figure~\ref{fig:ph_rendering} compares frozen GUAVA, an Adapter, and our tunneling decoder through synchronized facial crops. The 512-resolution baseline is enlarged for display. Both adapted outputs are generated at 1024 resolution, and the visualization presents the combined effect of adaptation and native-resolution decoding. All trainable variants use the same frames, Gaussian features, targets, optimizer, and update budget. On held-out frames, Adapter reduces mean RGB L1 by 13.5\%, while tunneling reduces it by 24.2\% relative to the frozen baseline.

\subsection{Ablation III: Test-Time Body Primitives}
Under a fixed 249-step adaptation budget and chronological 80\%/20\% training--evaluation split, Table~\ref{tab:body_capacity} shows that increasing $K$ from 512 to 1,024 reduces refined RGB L1 from 0.0599 to 0.0575 and boundary L1 from 0.0282 to 0.0270.

Figure~\ref{fig:body_sq} compares frozen GUAVA with identity-specific Body SQ under the same frame, pose, and camera; the learned layer locally adapts the carrier around the torso and upper arms.

Reference overlays locate the learned patches, carrier views expose their geometric support, and RGB errors use the same $[0,0.08]$ range with darker colors indicating larger error, separating surface alignment from its effect on rendering. Figure~\ref{fig:body_sq} highlights the resulting geometric fit, and Table~\ref{tab:body_capacity} confirms simultaneous gains in RGB fidelity, boundary alignment, and structural similarity.

% Retained quantitative discussion/table for a later revision.
% \subsection{Ablation II: Test-Time Detail Tunneling}
% The detail study contrasts frozen decoding with interpolation, LoRA, Adapter, a zero-initialized residual control, and tunneling (Table~\ref{tab:detail_ablation}). Shared avatars, geometry, video frames, losses, resolution, and training budgets isolate the adaptation mechanism. Parameter-matched controls distinguish architectural effects from capacity. The residual control replaces the independent softmax path while retaining the surrounding adaptation structure.
%
% Identical native-resolution crops reveal local texture and boundary fidelity. Consecutive frames expose ghosting and flicker. Figure~\ref{fig:ph_rendering} illustrates the face and hand detail comparison used in SignSplat.
% \input{tables/main_results/detail_ablation}
%
%

% Reserve the current left column for the body discussion.
\suppressfloats[t]
\begin{table}[t]
\centering\small
\setlength{\tabcolsep}{5pt}
\resizebox{\linewidth}{!}{%
\begin{tabular}{l|c|r}
\hline
Stage & Output & Throughput $\uparrow$\\\hline
Motion synthesis & SMPLest-X~\cite{yin2025smplestxultimatescalingexpressive} motion & 198.7 FPS\\
Primitive deformation & Face and body primitives & 320.9 FPS\\
Avatar rendering & HD RGB & 52.0 FPS\\
Detail refinement & 4K RGB & 47.6 FPS\\
Resident pipeline & Text $\rightarrow$ HD RGB & 25.3 FPS\\\hline
\end{tabular}}
\caption{\textbf{Online efficiency.} Component rows report warmed batch-one throughput on one Quadro RTX 8000. The resident row reports measured wall-clock throughput over 784 frames, including composition, motion synthesis, HD rendering, encoding, and concatenation.}
\label{tab:preparation_cost}
\end{table}

\begin{table}[t]
\centering
\small
\setlength\tabcolsep{7pt}
\renewcommand{\arraystretch}{1.12}
\resizebox{0.75\linewidth}{!}{%
\begin{tabular}{lcc}
\toprule
\textbf{Configuration} & \textbf{B-T BLEU-4}$\uparrow$ & \textbf{ROUGE-L}$\uparrow$ \\
\midrule
Enhancement disabled & 25.6450 & 59.4771 \\
\textbf{Enhancement enabled} & \textbf{26.8897} & \textbf{61.7298} \\
\bottomrule
\end{tabular}%
}
\caption{\textbf{Semantic preservation under visual enhancement.} Both configurations render identical motions. The enabled configuration activates shared face primitives, test-time body primitives, and detail tunneling. Signing poses are re-extracted from the final RGB videos and translated by the same frozen model.}
\label{tab:back_translation}
\end{table}

\begin{figure*}[t]
\centering
\includegraphics[width=\textwidth]{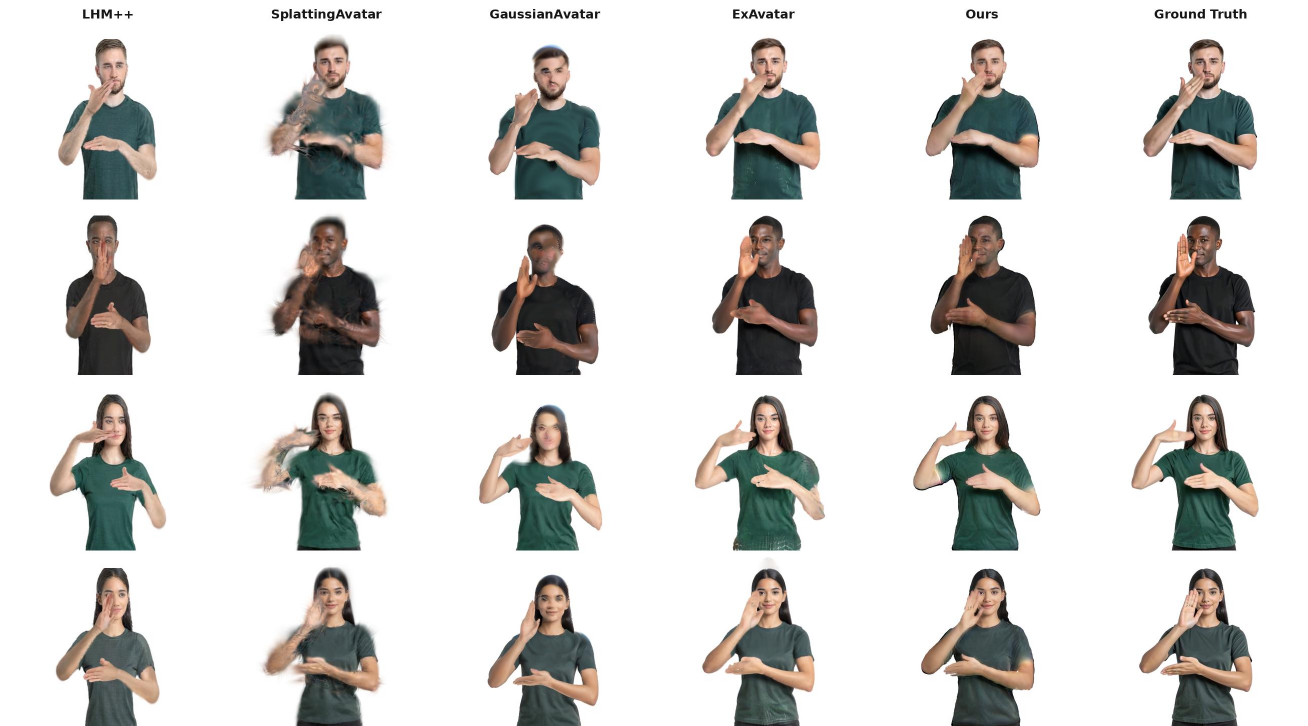}
\caption{\textbf{Qualitative rendering comparison.} Each column corresponds to one method: LHM++, SplattingAvatar, GaussianAvatar, ExAvatar, Ours, and Ground Truth. Each row shows the same held-out identity and signing pose across methods; the comparison focuses on facial detail, finger structure, and garment boundaries.}
\label{fig:rendering_comparison}
\end{figure*}

\subsection{Comparison with Existing Methods}

\noindent\textbf{Online efficiency.} Table~\ref{tab:preparation_cost} reports warmed batch-one component throughput on a single Quadro RTX 8000. Motion synthesis and primitive deformation run at 198.7 and 320.9 FPS, while HD avatar rendering and 4K detail refinement reach 52.0 and 47.6 FPS. In the resident pipeline, stages overlap across streaming chunks. A warmed 784-frame, three-chunk text-to-HD-avatar run completes in 31.0 seconds wall time (25.3 generated FPS), including composition, motion synthesis, streamed rendering, encoding, and chunk concatenation. Body and detail adaptation are performed once for each new avatar and baked into the deployed model; subsequent sequences run directly from the baked parameters.

\noindent\textbf{Back-translation evaluation.} Table~\ref{tab:back_translation} tests whether the proposed visual enhancement components alter the linguistic content of the output. Both configurations render the same motion sequence; the full configuration additionally activates shared face primitives, test-time body primitives, and detail tunneling. We re-extract signing poses from each rendered video and translate them with the same frozen back-translation model. Activating all three components raises BLEU-4 from 25.6450 to 26.8897 (4.9\%) and ROUGE-L from 59.4771 to 61.7298 (3.8\%). The complete enhancement stack therefore improves geometric and rendering fidelity while fully preserving the underlying signs.

\noindent\textbf{Rendering fidelity.} Table~\ref{tab:rendering_fidelity} first compares our system with four reproducible open-source avatar methods over all video frames in the complete evaluation set. We compute PSNR inside the shared ground-truth foreground mask and SSIM/LPIPS within the same ground-truth person bounding box, then macro-average reference videos. The second block evaluates our identity-adapted renderer against frozen GUAVA using matched cached features and crops sampled directly from held-out 4K targets.

The second block measures identity adaptation by comparing frozen GUAVA and our complete renderer on the same held-out subset from each of ten reference videos. Both methods receive identical cached Gaussian features, crop coordinates, and $768{\times}768$ crops sampled directly from the held-out 4K frames. Macro-averaged over identities, Stable Signer V3 raises PSNR from 19.418 to 21.790\,dB and SSIM from 0.833 to 0.860, while reducing LPIPS from 0.245 to 0.226. This corresponds to a 2.372\,dB PSNR gain and an 8.0\% relative LPIPS reduction over the frozen renderer.

\subsection{Qualitative Rendering Comparison}
Figure~\ref{fig:rendering_comparison} compares methods column by column on the same sign-language sequence. The four published baselines are followed by our method and Ground Truth. The shared framing makes facial detail, finger articulation, and garment boundaries directly comparable across columns. Across identities and poses, our method preserves facial character, articulated fingers, silhouettes, and garment boundaries more consistently than the reproduced baselines. Faces remain stable. Hands stay distinct during articulation. Garment silhouettes follow motion cleanly.

% The first block is measured on four provided input reference videos.
% The baselines are measured by make/table4_evaluation/evaluate_avatar_baselines.py.
% The first block, including our row, is measured by make/table4_evaluation/evaluate_avatar_baselines.py.
% The second block is measured by make/table4_evaluation/evaluate_native4k.py.
% Values must only be compared within a protocol block.
\begin{table}[t]
\centering\small
\setlength{\tabcolsep}{4pt}
\resizebox{0.78\columnwidth}{!}{%
\begin{tabular}{l|rrr}
\hline
Method & PSNR $\uparrow$ & SSIM $\uparrow$ & LPIPS $\downarrow$\\\hline
\multicolumn{4}{l}{\textit{Complete evaluation-set video evaluation}}\\
LHM++~\cite{qiu2026lhmefficientlargehuman} & 8.987 & 0.646 & 0.576\\
SplattingAvatar~\cite{SplattingAvatar:CVPR2024} & 18.396 & 0.764 & 0.348\\
GaussianAvatar~\cite{hu2024gaussianavatar} & 10.246 & 0.684 & 0.460\\
ExAvatar~\cite{moon2024exavatar} & 18.141 & 0.775 & 0.291\\
\textbf{Stable Signer V3} & \textbf{19.415} & \textbf{0.787} & \textbf{0.276}\\\hline
\multicolumn{4}{l}{\textit{Matched held-out reference-video evaluation (10 identities)}}\\
Frozen GUAVA~\cite{Zhang_2025_ICCV} & 19.418 & 0.833 & 0.245\\
\textbf{Stable Signer V3} & \textbf{21.790} & \textbf{0.860} & \textbf{0.226}\\\hline
\end{tabular}%
}
\caption{\textbf{Rendering fidelity.} Macro-averaged results under two matched protocols: all video frames in the complete evaluation set (top) and crops sampled directly from held-out 4K targets for ten identities (bottom). PSNR uses GT foreground; SSIM and LPIPS use the shared GT person crop.}
\label{tab:rendering_fidelity}
\vspace{-6pt}
\end{table}

\section{Conclusion}
\label{sec:conclusion}
We presented Stable Signer V3, a sign language production framework combining shared facial primitives, test-time identity-specific body primitives, and a tunneling decoder for high-resolution appearance. Together, they unify expressive control and personalized rendering through a reusable deformable representation. Reference-video adaptation is performed once; stored identity parameters then drive later signing. On held-out evaluations, the modules reduce head error by 71.4\%, body-boundary error by 35.1\% across ten identities, and high-resolution RGB error by 24.2\%, while retaining interactive inference. Together, these gains enable stable, expressive, high-fidelity sign language avatars at up to 4K resolution.
% Future work will broaden signer and motion coverage and include evaluation by proficient sign-language users.

\clearpage
{
    \small
    \bibliographystyle{ieeenat_fullname}
    \bibliography{ref/main,ref/hugediff_source,ref/hugediff_baselines,ref/signsplat_source,ref/sds,ref/meta,ref/llm,ref/rl,ref/slt,ref/how2sign}
}

% Remove this line when exporting a main-paper-only submission PDF.
\clearpage
\setcounter{page}{1}
\maketitlesupplementary

\section{Evaluation Protocols}
\label{sec:supp_protocols}

We expand the evaluation protocols, component analyses, and deployment measurements presented in the main paper. Supervision and evaluation occupy temporally disjoint segments, and checkpoint selection uses the supervision segment exclusively. Paired methods share motion, camera, background, crop, and target data. The controlled face and body studies use fixed optimization budgets.

\subsection{Face primitive evaluation}
We evaluate facial deformation against registered target geometry. The first 80\% trains the shared primitive controller, and the final 20\% measures generalization to later motion. Full-expression FLAME and FLAME+SQ share all fifty expression coefficients together with global pose, jaw, eye, and eyelid articulation. We compute Euclidean error in the common registered coordinates.

The controller contains $K{=}1{,}024$ local primitives, four-neighbor blending, and a compact residual MLP with 2.42M trainable parameters. We optimize it for 1,200 iterations using batch size 8 and a learning rate of $2\!\times\!10^{-3}$. Table~\ref{tab:supp_face_regions} resolves the aggregate head error into semantic regions. Face primitives reduce error across the lips, eyes, forehead, and nose, confirming that the improvement spans the expressive facial surface.

\begin{table}[t]
\centering
\small
\resizebox{\linewidth}{!}{%
\begin{tabular}{lrrr}
\toprule
Region & FLAME & FLAME+SQ & Reduction \\
 & (mm)$\downarrow$ & (mm)$\downarrow$ & (\%)$\uparrow$ \\
\midrule
Full head & 1.267 & \textbf{0.362} & 71.4 \\
Face      & 1.492 & \textbf{0.518} & 65.3 \\
Lips      & 1.739 & \textbf{0.990} & 43.1 \\
Left eye  & 1.029 & \textbf{0.587} & 43.0 \\
Right eye & 1.483 & \textbf{0.626} & 57.8 \\
Forehead  & 1.047 & \textbf{0.339} & 67.6 \\
Nose      & 1.009 & \textbf{0.477} & 52.7 \\
\bottomrule
\end{tabular}}
\caption{\textbf{Region-wise face reconstruction error.} Mean Euclidean point error on the held-out 20\% of the tracked sequence. Both methods use identical articulation and coordinates.}
\label{tab:supp_face_regions}
\end{table}

\subsection{Body adaptation evaluation}
The body capacity study uses a chronological 80\%/20\% split of one identity-specific reference video. We uniformly sample a fixed evaluation subset and train the $K{=}512$ and $K{=}1{,}024$ models for the same 249-step budget. Both configurations share four-neighbor blending, optimizer, supervision, and loss weights. Their objective combines raw and refined RGB reconstruction, silhouette and boundary alignment, image gradients, SSIM, anchor regularization, and spatial smoothness.

Table~\ref{tab:supp_body_full} further evaluates the final $K{=}1{,}024$ checkpoint over the complete held-out segment. With the carrier and renderer fixed, continued adaptation improves appearance and boundary alignment. Its 1.25-cm mean displacement shows that compact local corrections are sufficient to capture identity-specific body geometry.

\begin{table}[t]
\centering
\small
\resizebox{\linewidth}{!}{%
\begin{tabular}{lrrrr}
\toprule
Model & RGB L1$\downarrow$ & Boundary L1$\downarrow$ & SSIM$\uparrow$ & Disp. (m)$\downarrow$ \\
\midrule
Frozen carrier & 0.0649 & 0.0350 & 0.9038 & -- \\
$K{=}512$, matched budget & 0.0599 & 0.0282 & 0.9076 & 0.0111 \\
$K{=}1{,}024$, matched budget & 0.0575 & 0.0270 & 0.9096 & 0.0111 \\
$K{=}1{,}024$, final checkpoint & \textbf{0.0562} & \textbf{0.0230} & \textbf{0.9115} & 0.0125 \\
\bottomrule
\end{tabular}}
\caption{\textbf{Body adaptation over the held-out segment.} The first three rows reproduce the controlled capacity study; the final row reports the converged model on the complete held-out segment.}
\label{tab:supp_body_full}
\end{table}

We then fit the same $K{=}1{,}024$ configuration independently to ten characters. Table~\ref{tab:supp_body_identities} reports macro-averages over their complete held-out segments. Body primitives improve refined RGB L1, boundary L1, and SSIM for every character, demonstrating consistent per-character adaptation across the full identity set.

\begin{table}[t]
\centering
\small
\begin{tabular}{lrrr}
\toprule
Method & RGB L1$\downarrow$ & Boundary L1$\downarrow$ & SSIM$\uparrow$ \\
\midrule
Frozen carrier & 0.0661 & 0.0308 & 0.9086 \\
Body primitives & \textbf{0.0551} & \textbf{0.0200} & \textbf{0.9155} \\
\bottomrule
\end{tabular}
\caption{\textbf{Ten-identity body adaptation.} Macro-average over temporally held-out segments. The same model configuration and evaluation code are used for all identities.}
\label{tab:supp_body_identities}
\end{table}

\section{Detail Tunneling Analysis}
\label{sec:supp_tunnel}

Native-resolution reference video supervises the detail module with the Gaussian predictor and StyleUNet~\cite{Wang_2021_CVPR} backbone fixed. Character preparation bakes the learned identity parameters into the decoder, and online sequences run directly through this personalized model. The design concentrates reference evidence into a one-time adaptation stage and preserves a single-stream inference path.

Table~\ref{tab:supp_tunnel} compares the trainable alternatives under a shared 900-step budget. LoRA and Adapter both improve the frozen decoder; the independent late Softmax pathway achieves the largest reconstruction gain. Increasing memory from 64 to 128 slots adds only 0.08 dB. We therefore adopt rank-8 tunneling with 64 slots as the default operating point.

\begin{table}[t]
\centering
\small
\resizebox{\linewidth}{!}{%
\begin{tabular}{lrr}
\toprule
Method & RGB L1$\downarrow$ & PSNR$\uparrow$ \\
\midrule
Frozen decoder & 0.0260 & 22.16 \\
LoRA ($r{=}32$) & 0.0234 & 22.62 \\
Adapter ($r{=}32$) & 0.0225 & 22.90 \\
Tunneling ($r{=}8$, 64 slots) & \textbf{0.0197} & 23.77 \\
Tunneling ($r{=}8$, 128 slots) & 0.0196 & \textbf{23.85} \\
\bottomrule
\end{tabular}}
\caption{\textbf{Equal-budget detail adaptation.} Every trainable method is evaluated after 900 update steps on the same identity, held-out frames, Gaussian features, and native-resolution targets.}
\label{tab:supp_tunnel}
\end{table}

At convergence, the selected 64-slot model reaches 0.0181 RGB L1 and 24.43 dB PSNR; the 128-slot model reaches 0.0178 and 24.48 dB. This 0.05-dB gap shows that 64 slots capture nearly all of the detail gain, supporting the selected default.

\section{Per-Identity 4K Results}
\label{sec:supp_identity}

Table~\ref{tab:supp_identity_4k} reports native-4K results for all ten identities. Each renderer receives identical Gaussian features and deterministic aligned $768{\times}768$ crops from held-out 4K targets. We compute each identity score first and then macro-average, giving every character equal weight.

\begin{center}
\scriptsize
\setlength{\tabcolsep}{2.5pt}
\resizebox{\linewidth}{!}{%
\begin{tabular}{c|rr|rr|rr}
\toprule
& \multicolumn{2}{c|}{PSNR$\uparrow$} & \multicolumn{2}{c|}{SSIM$\uparrow$} & \multicolumn{2}{c}{LPIPS$\downarrow$} \\
Identity & GUAVA & Ours & GUAVA & Ours & GUAVA & Ours \\
\midrule
S01 & 20.146 & \textbf{22.956} & 0.8487 & \textbf{0.8715} & 0.2243 & \textbf{0.1988} \\
S02 & 18.903 & \textbf{21.538} & 0.8452 & \textbf{0.8719} & 0.2123 & \textbf{0.1900} \\
S03 & 20.324 & \textbf{23.713} & 0.8316 & \textbf{0.8715} & 0.2189 & \textbf{0.2004} \\
S04 & 19.554 & \textbf{22.304} & 0.8105 & \textbf{0.8556} & 0.2373 & \textbf{0.2220} \\
S05 & 19.213 & \textbf{21.766} & 0.8460 & \textbf{0.8689} & 0.2294 & \textbf{0.2049} \\
S06 & 20.653 & \textbf{22.549} & 0.8483 & \textbf{0.8670} & 0.2453 & \textbf{0.2220} \\
S07 & 18.199 & \textbf{19.631} & 0.8240 & \textbf{0.8387} & 0.2862 & \textbf{0.2748} \\
S08 & 18.434 & \textbf{20.675} & 0.7829 & \textbf{0.8195} & 0.2873 & \textbf{0.2744} \\
S09 & 19.557 & \textbf{21.838} & 0.8421 & \textbf{0.8664} & 0.2485 & \textbf{0.2273} \\
S10 & 19.198 & \textbf{20.928} & 0.8527 & \textbf{0.8687} & 0.2645 & \textbf{0.2440} \\
\midrule
Mean & 19.418 & \textbf{21.790} & 0.8332 & \textbf{0.8600} & 0.2454 & \textbf{0.2259} \\
\bottomrule
\end{tabular}}
\captionof{table}{\textbf{Per-identity native-4K rendering fidelity.} The adapted renderer improves all three metrics for every evaluated identity. Identity labels are anonymized and fixed across columns.}
\label{tab:supp_identity_4k}
\end{center}

\section{Runtime and Deployment}
\label{sec:supp_runtime}

\paragraph{Stage throughput and latency.}
On one Quadro RTX 8000, warmed batch-one throughput reaches 198.7 FPS for motion synthesis, 320.9 FPS for primitive deformation, 52.0 FPS for avatar rendering, and 47.6 FPS for 4K detail refinement. The resident pipeline overlaps stages across chunks. A warmed 784-frame, three-chunk text-to-HD-avatar request completes in 31.0 seconds, yielding 25.3 FPS across composition, diffusion, streaming render, encoding, and concatenation. This wall-clock measurement captures the concurrency of the deployed pipeline.

\paragraph{One-time character preparation.}
The shared face controller transfers directly across characters. Character preparation fits the body and detail modules once and stores them with the avatar. Each final $K{=}1{,}024$ body model contains 2.43M trainable parameters and occupies 9.6 MB. Across ten identities, body adaptation averages 40.2 minutes on one Quadro RTX 8000, with a 22.6--53.5 minute range driven mainly by clip length. Online inference loads the resulting character package directly.

\paragraph{Meaning of 4D deformable primitives.}
In our formulation, \emph{4D} indexes primitive transformations by pose and time throughout a motion sequence. The primitives supply local deformable support for the surface-bound Gaussian avatar, and Gaussian splatting rasterizes the resulting features. This factorization adds expressive local control to the pretrained parametric carrier while preserving its avatar representation and rendering path.

\section{Interactive User Guide}
\label{sec:supp_user_guide}

\begin{figure*}[t]
\centering
\includegraphics[width=0.96\textwidth]{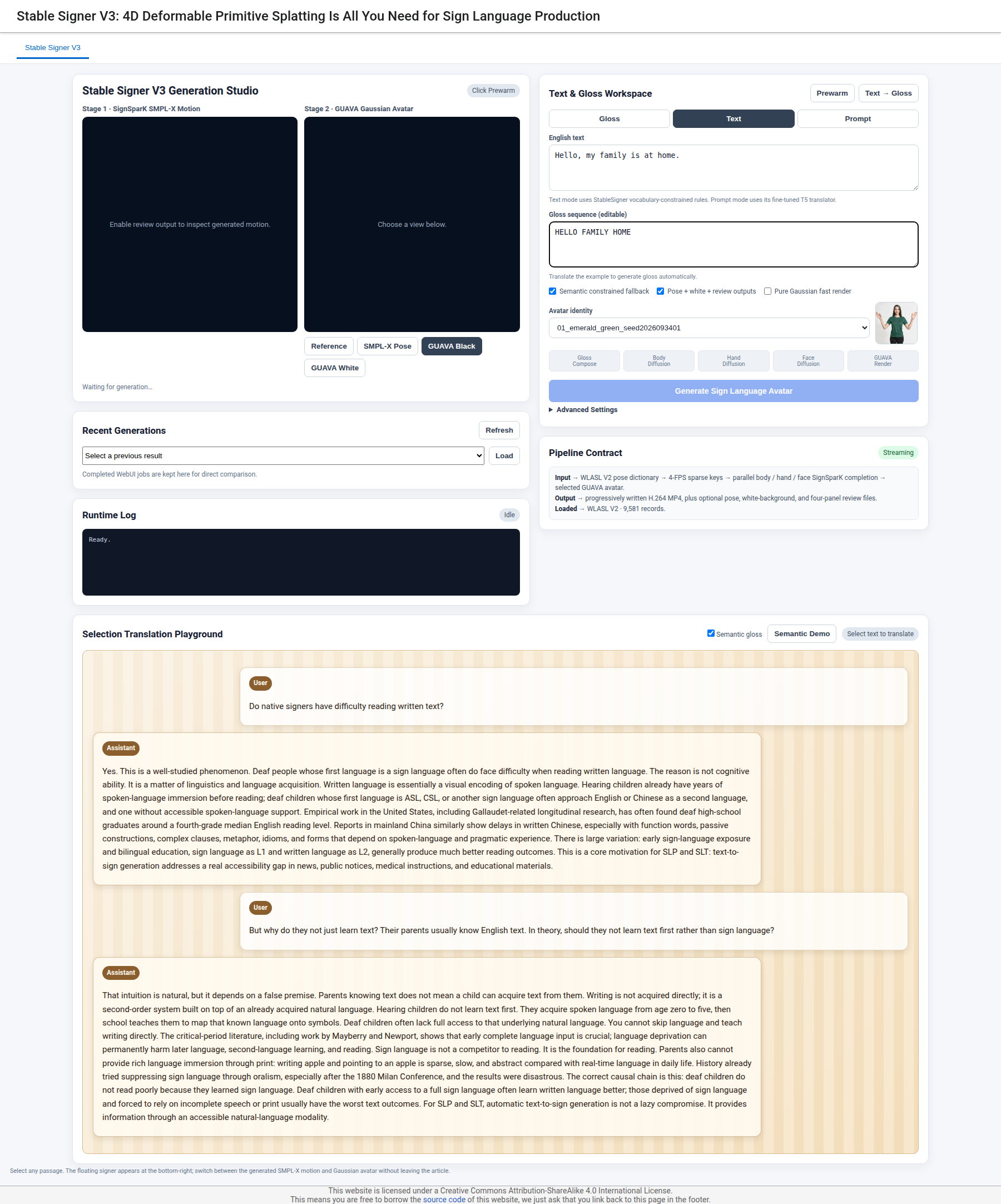}
\caption{\textbf{Stable Signer V3 interface overview.} The complete interface presents text or gloss input, signer identity, staged motion and avatar outputs, execution state, and reusable results within one workspace.}
\label{fig:supp_webui_overview}
\end{figure*}

\begin{figure*}[t]
\centering
\begin{subfigure}[t]{0.492\textwidth}
  \includegraphics[width=\linewidth]{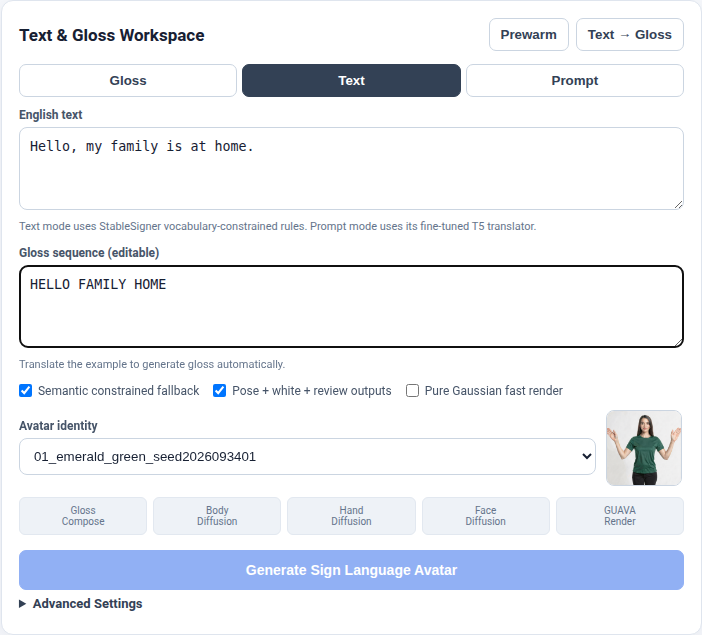}
  \caption{Enter text or gloss and select the signer identity.}
\end{subfigure}\hfill
\begin{subfigure}[t]{0.492\textwidth}
  \includegraphics[width=\linewidth]{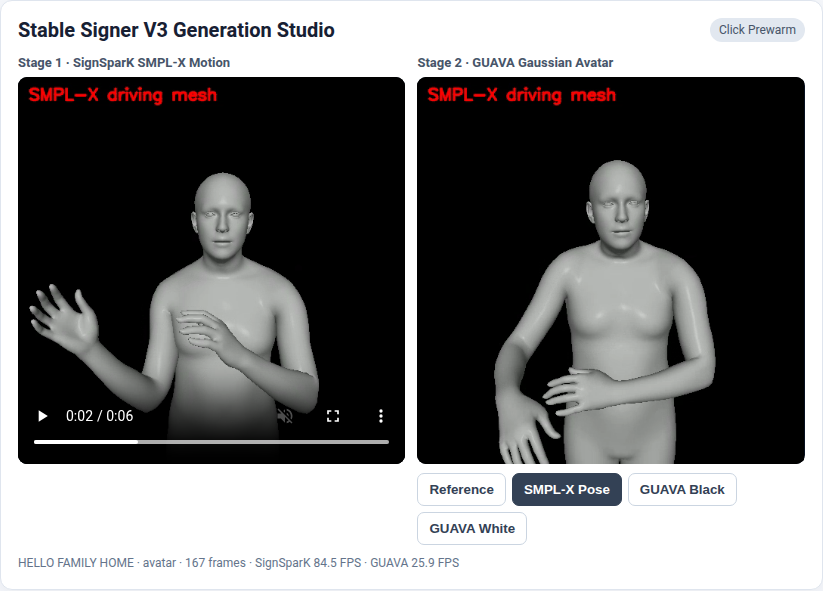}
  \caption{Inspect the synthesized SMPL-X signing motion.}
\end{subfigure}

\vspace{2pt}
\begin{subfigure}[t]{0.492\textwidth}
  \includegraphics[width=\linewidth]{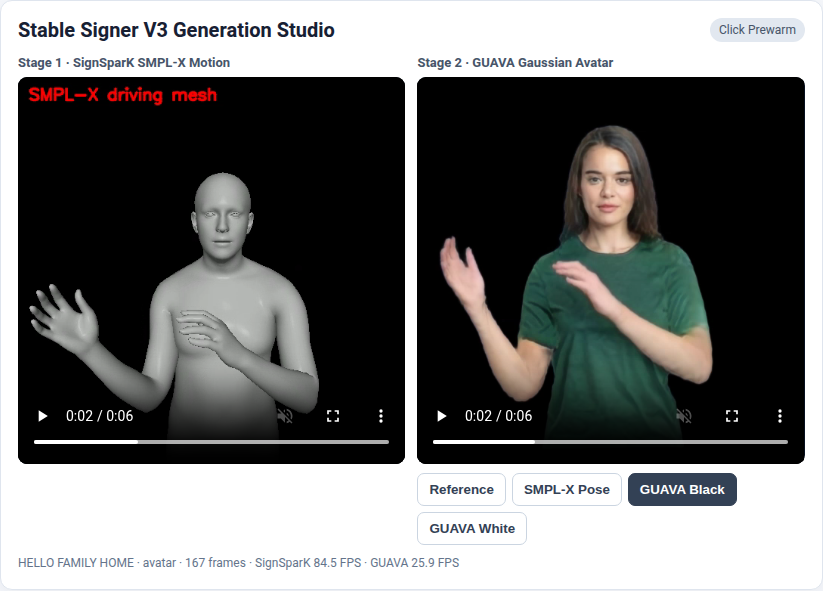}
  \caption{Review the Gaussian avatar rendering.}
\end{subfigure}\hfill
\begin{subfigure}[t]{0.492\textwidth}
  \includegraphics[width=\linewidth]{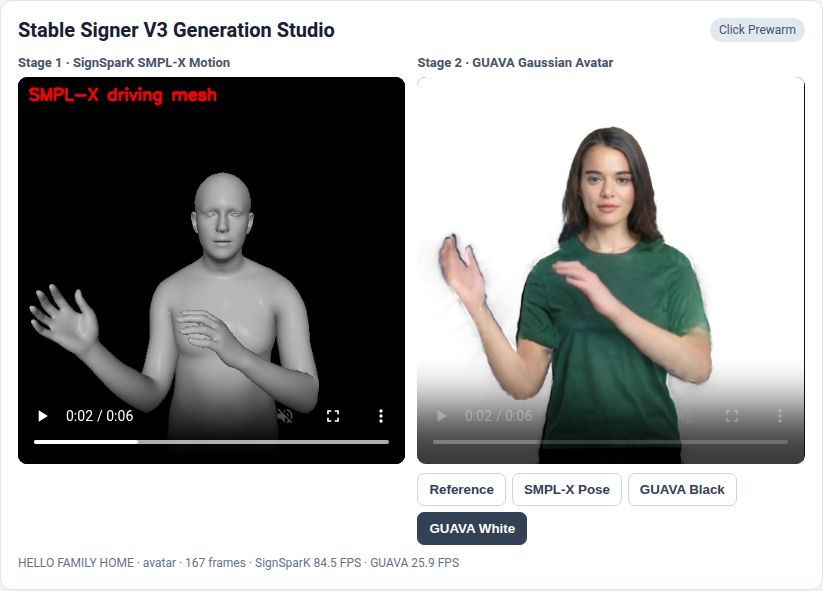}
  \caption{Switch to the white-background final output.}
\end{subfigure}
\caption{\textbf{Stable Signer V3 interactive generation workflow.} The interface exposes input translation, identity selection, stage-level progress, motion inspection, and final avatar outputs in one view. The example loads the same completed generation across all output modes.}
\label{fig:supp_webui_guide}
\end{figure*}

Figures~\ref{fig:supp_webui_overview} and~\ref{fig:supp_webui_guide} demonstrate the complete generation workflow in the Stable Signer V3 interface. A user can begin with English text, an instruction-conditioned prompt, or an editable gloss sequence, then select an avatar identity from its reference thumbnail. After a single prewarm operation, \emph{Generate Sign Language Avatar} executes gloss composition, body, hand, and face motion synthesis, followed by Gaussian-avatar rendering. The interface streams intermediate results and exposes the generated SMPL-X motion for direct inspection. The same job can then be viewed as the Gaussian avatar on black or white backgrounds, while completed requests remain available under \emph{Recent Generations}. This interface turns the full text-to-4K-avatar system into a short, inspectable workflow and requires no manual exchange of intermediate files.

\begin{figure*}[t]
\centering
\includegraphics[width=0.72\textwidth]{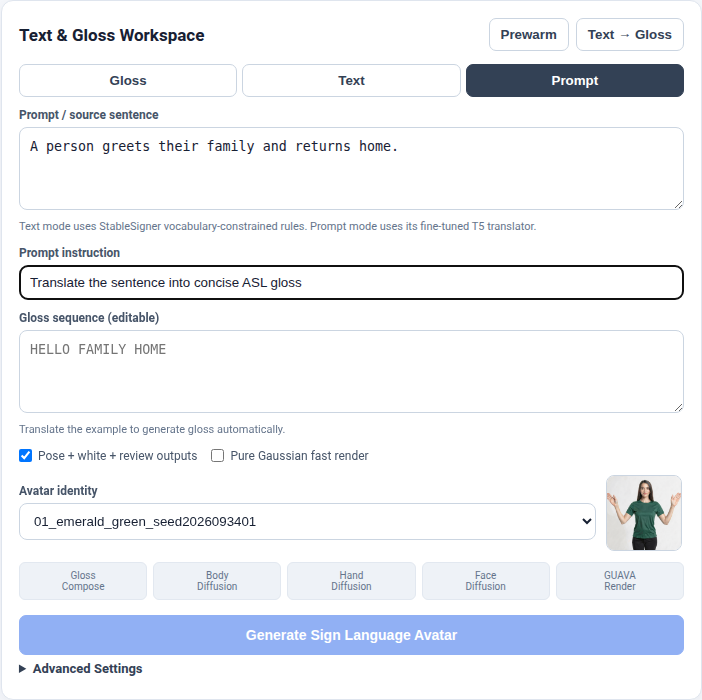}
\caption{\textbf{Instruction-conditioned generation.} Prompt input, editable gloss, semantic fallback, signer identity, and the complete generation command remain visible within one uninterrupted workspace.}
\label{fig:supp_webui_prompt}
\end{figure*}

\begin{figure*}[t]
\centering
\begin{subfigure}[t]{0.492\textwidth}
  \includegraphics[width=\linewidth]{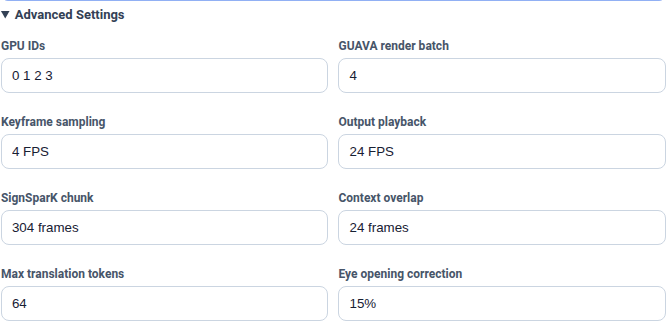}
  \caption{Runtime controls for devices, batching, temporal sampling, and sequence overlap.}
\end{subfigure}\hfill
\begin{subfigure}[t]{0.492\textwidth}
  \includegraphics[width=\linewidth]{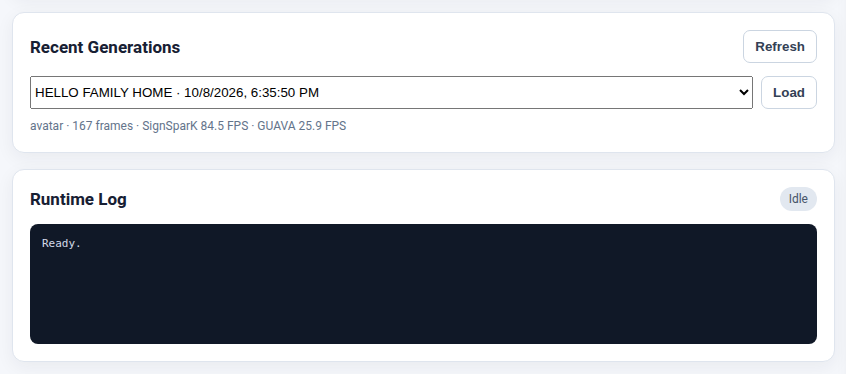}
  \caption{Persistent generation history, measured throughput, and asynchronous runtime logs.}
\end{subfigure}
\caption{\textbf{Execution and inspection controls.} Stable Signer V3 exposes reproducible runtime settings and keeps completed generations available for immediate comparison.}
\label{fig:supp_webui_features}
\end{figure*}

\begin{figure*}[t]
\centering
\begin{subfigure}[t]{0.492\textwidth}
  \includegraphics[width=\linewidth]{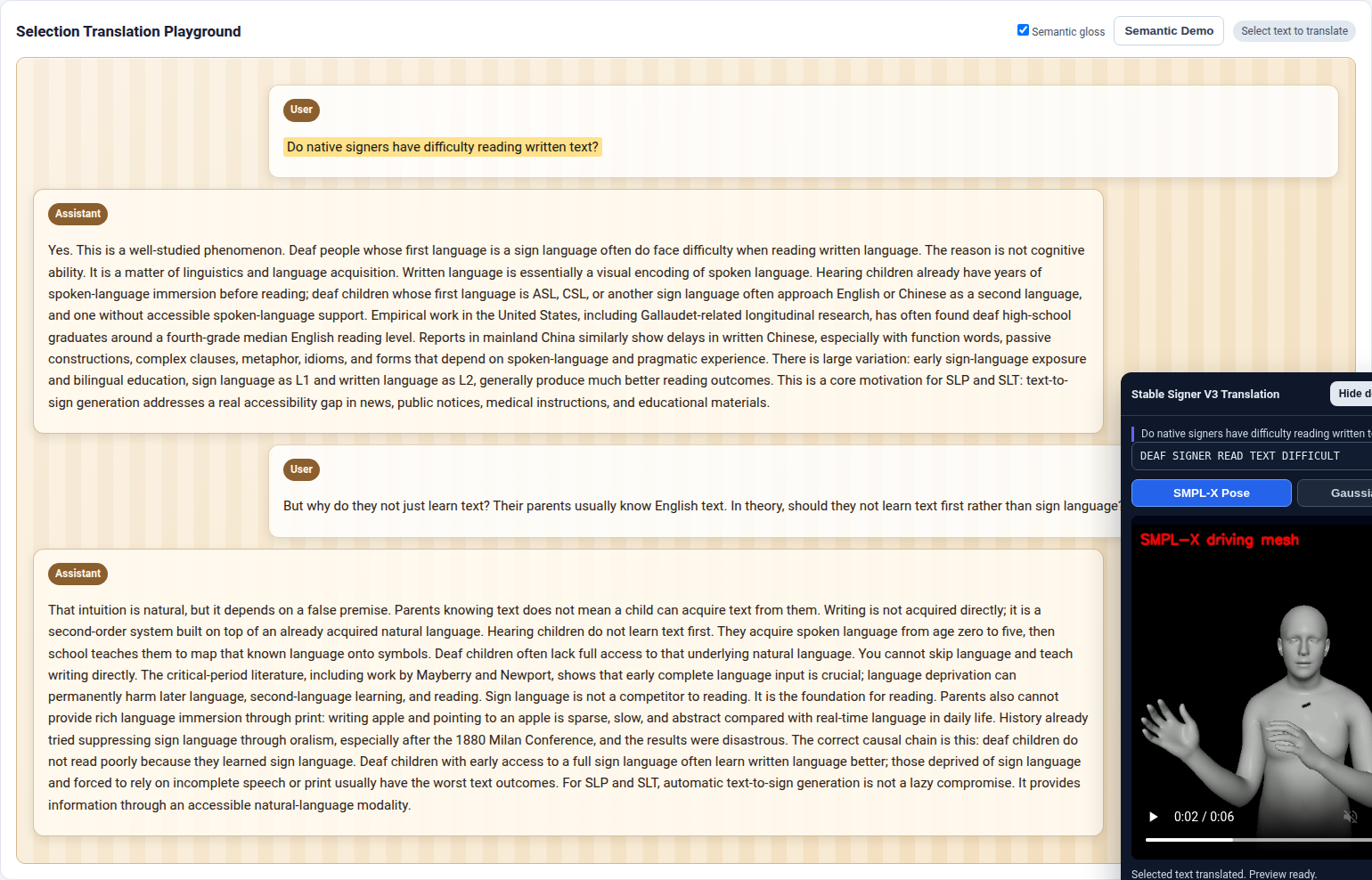}
  \caption{Selected text opens a bottom-right signer with SMPL-X motion playback.}
\end{subfigure}\hfill
\begin{subfigure}[t]{0.492\textwidth}
  \includegraphics[width=\linewidth]{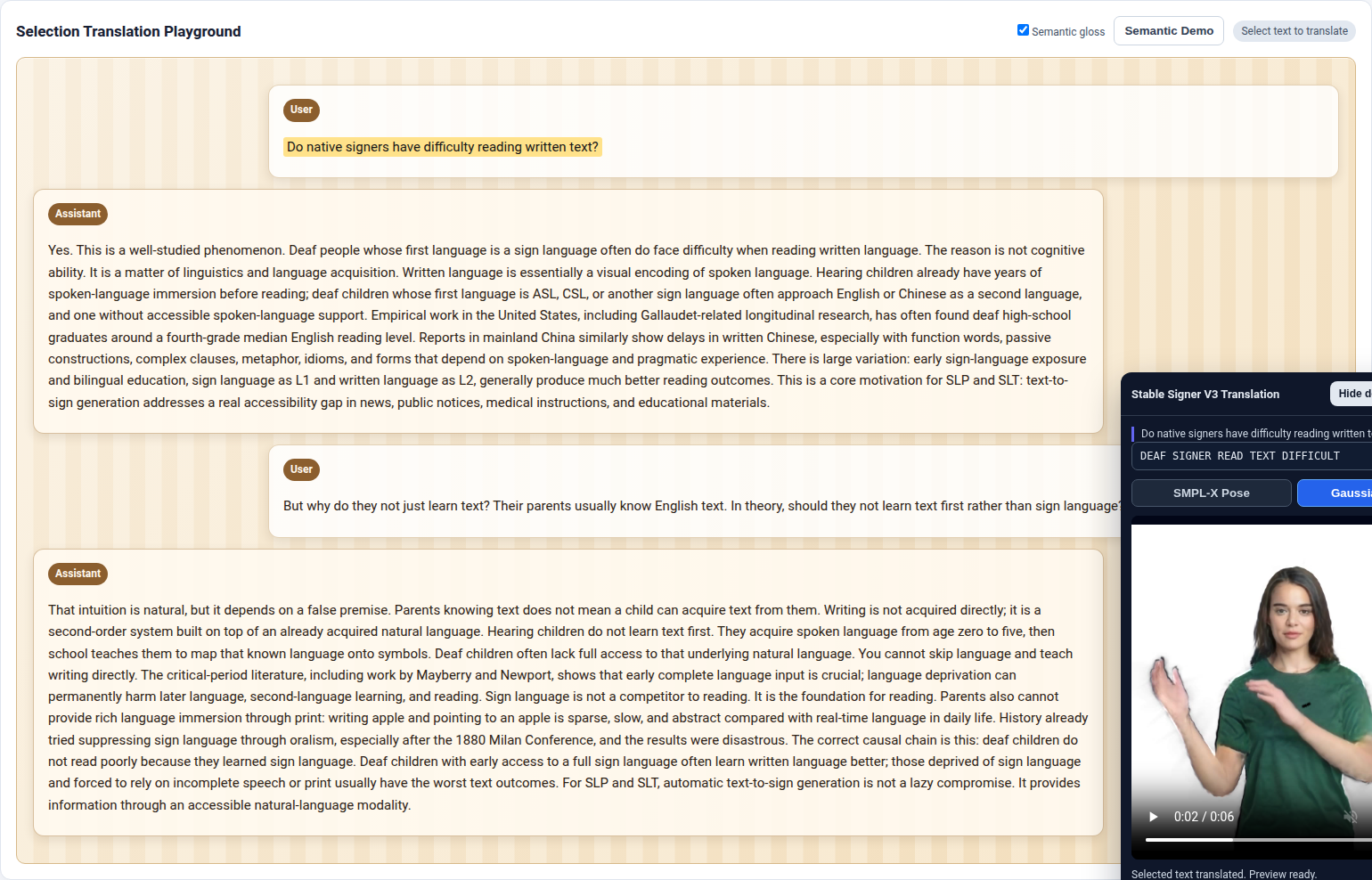}
  \caption{The same selection switches directly to the rendered Gaussian avatar.}
\end{subfigure}
\caption{\textbf{Contextual text-to-sign access.} Highlighting a passage invokes the floating signer in place and provides both its 3D motion and photorealistic avatar views without leaving the document.}
\label{fig:supp_webui_selection}
\end{figure*}

\end{document}